\documentclass[letterpaper,10pt,conference]{ieeeconf}
\IEEEoverridecommandlockouts
\usepackage[T1]{fontenc}
\usepackage{mathptmx}
\usepackage{amsmath,amssymb,bm}
\usepackage{graphicx}
\usepackage{booktabs,tabularx}
\usepackage{cite,url,xcolor}

\graphicspath{{./}{figure/}{figures/}}

\title{Selective Commitment for Language-Guided Object Retrieval under Partial Observability}

\author{
Wonhee Koh$^{1}$,
Sushil Samuel Dinesh$^{2}$,
Hansol Ko$^{1}$, Shinkyu Park$^{2}$,
and Eungjoo Lee$^{1}$%
\thanks{$^{1}$Wonhee Koh, Hansol Ko, and Eungjoo Lee are with
the School of Electrical, Computing, and Software Engineering,
University of Arizona, Tucson, AZ 85721, USA.
\texttt{\{wonheekoh,kohanasol,eungjoolee\}@arizona.edu}}%
\thanks{$^{2}$Sushil Samuel Dinesh and Shinkyu Park are with
the Department of Electrical and Computer Engineering,
King Abdullah University of Science and Technology (KAUST),
Thuwal 23955, Saudi Arabia.
\texttt{\{sushilsamuel.dinesh,shinkyu.park\}@kaust.edu.sa}}%
}

\IEEEaftertitletext{%
  \vspace{-0.35\baselineskip}%
  \centering
  \includegraphics[width=\textwidth,keepaspectratio]{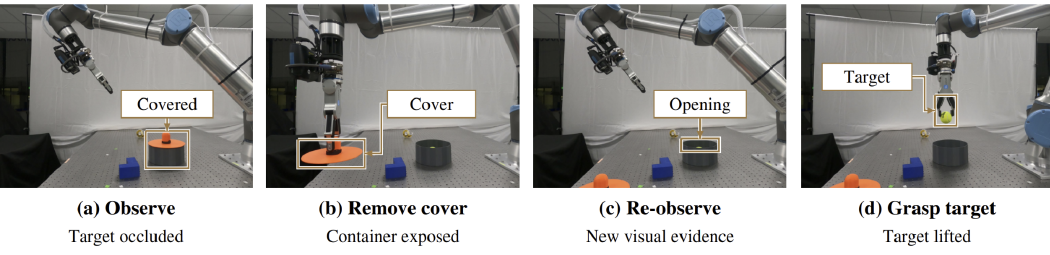}\\[-0.35ex]
  \refstepcounter{figure}\label{fig:real_robot_closed_loop}%
  {\normalfont\footnotesize Fig.~\thefigure. Object retrieval on the physical robot. The robot re-observes after cover removal and before grasping. Boxes highlight the relevant regions.}\\[0.55\baselineskip]%
}

\begin{document}
\bstctlcite{IEEEReferenceControl}
\maketitle

\begin{abstract}
Language-guided object retrieval under partial observability requires deciding whether to gather more evidence, interact with the scene, grasp a candidate, or abstain. We present a closed-loop framework that coordinates these decisions for retrieving a target specified in relation to a reference container. The framework maintains a persistent joint belief over target identity, container relation, and presence through tracked-object, unobserved-target, and target-absent hypotheses. View-conditioned categorical VLM observations update this belief; conformal grasp eligibility and robot feasibility govern commitment, while finite-horizon belief-space planning selects information-gathering actions. Across five different scenarios, our proposed method succeeds in 19/25 simulation episodes versus 12/25 for the best-performing task-adapted baseline and is the only evaluated policy to achieve at least one success in each scenario. Ablations show that cross-view memory improves success under partial occlusion, while the full system does not consistently outperform simplified variants. Real-robot trials demonstrate closed-loop re-observation and autonomous recovery from injected grasp failures, while injected viewpoint failures end in false defer. Experimental results demonstrate the feasibility of coordinating evidence gathering and selective grasp commitment within a unified framework for retrieval under partial observability.
\end{abstract}

\section{Introduction}

Language-guided retrieval under partial observability requires preserving target identity across views and avoiding distractor grasps when evidence is ambiguous. Missing target evidence may reflect occlusion, viewpoint, perception error, or true absence. Fig.~\ref{fig:real_robot_closed_loop} illustrates the resulting closed-loop sequence on the physical robot. Unlike single-view recognition, retrieval requires the robot to decide  when to acquire more evidence and when to commit physically. For example, a target may be partially visible, hidden by a cover, confused with a similar object, or absent. A current-frame prediction alone cannot determine whether uncertainty should be resolved by changing viewpoint, interacting with the environment, or abstaining from grasping.

Closed-loop retrieval also requires maintaining hypotheses as the scene changes. Cover removal may reveal new objects, while viewpoint changes alter observable attributes and container relations. The state representation must therefore
preserve cross-view evidence, incorporate newly discovered objects, and retain an explicit alternative for target absence. The decision layer then maps this evolving uncertainty to information gathering, grasping, or abstention.

Belief-space planning addresses uncertainty and information gathering~\cite{curtis2024tampura}, active perception selects informative views~\cite{bajcsy2018revisiting,jauhri2023actpermoma}, and persistent representations support multi-view interaction~\cite{jiang2025roboexp,yan2025dovsg}. However, retrieval under target uncertainty requires coordinating these capabilities to decide whether to seek another observation, interact with the scene, grasp, or abstain while preserving target identity and container relation across views.

Our framework addresses this problem with a persistent belief over target identity, container relation, and presence. View-conditioned VLM observations update this belief across views and interactions, while selective commitment permits grasping only when semantic evidence and robot feasibility are sufficient; otherwise the robot gathers more information or abstains.

The main contributions of this study are:
\begin{enumerate}
\item An integrated closed-loop retrieval system evaluated in simulation and on a real robot, combining persistent object association, information gathering, and selective grasp commitment.

\item A joint belief over target identity, container relation, and presence that explicitly represents tracked-object hypotheses, an unobserved target, and target absence, with probability transferred from the unobserved-target hypothesis when new tracks are discovered.

\item View-conditioned categorical VLM observations for sequential belief updating, together with a conformal-based grasp eligibility rule, a defer action, and independent robot feasibility checks.
\end{enumerate}

\begin{figure*}[!t]
    \centering
    \includegraphics[width=\textwidth]{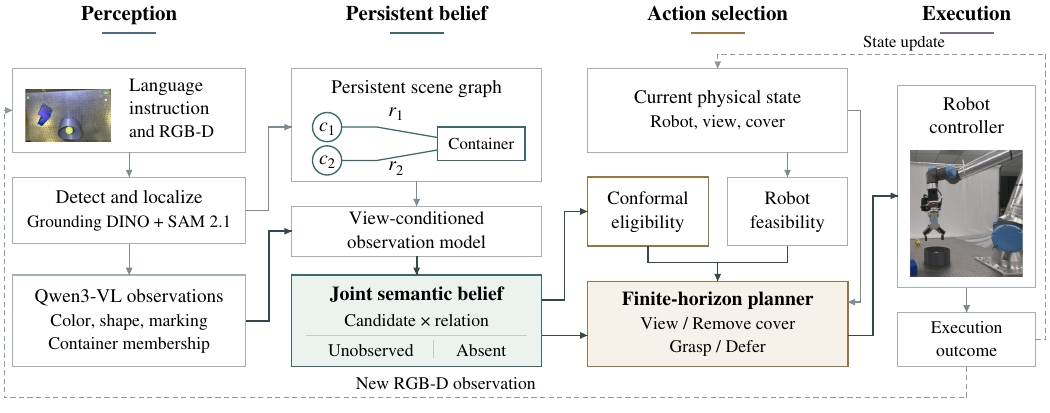}
    \caption{Closed-loop architecture. Categorical VLM observations update the joint belief using a view-conditioned observation model. The scene graph represents persistent object tracks $c_i$ and their container
relations $r_i$. A finite-horizon planner selects actions, with grasping subject to both conformal eligibility and robot feasibility. Execution outcomes update the physical state, and new RGB-D observations provide feedback.}
    \label{fig:system_architecture}
\end{figure*}

\section{Related Work}

\subsection{Language-Guided Manipulation and Active Perception}

VoxPoser~\cite{huang2023voxposer}, VLMPC~\cite{zhao2024vlmpc}, VLM-TAMP~\cite{yang2025vlmtamp}, and ReKep~\cite{huang2025rekep} connect language or VLM reasoning to manipulation. Active-perception methods explicitly incorporate information gathering: ActPerMoMa couples information gain with grasp reachability~\cite{jauhri2023actpermoma}, while SaPaVe integrates active camera control with vision-language-action manipulation ~\cite{liu2026sapave}. These works address complementary aspects of language-conditioned action generation and active sensing. Our focus is a retrieval setting in which evidence about target identity, container relation, and presence must be accumulated across observations and used to decide whether to gather more information, interact with the scene, grasp, or abstain.

\subsection{Persistent Memory, Belief-Space Planning, and Selective Commitment}

RoboEXP and DovSG maintain scene information across interactions or time~\cite{jiang2025roboexp,yan2025dovsg}. Our representation maintains persistent object identity, geometry, observation history, and relation to one reference container. TAMPURA addresses partially observable task-and-motion planning with information gathering and risk awareness~\cite{curtis2024tampura}. Seeing is Believing~\cite{zhao2025seeing} uses foundation-model uncertainty to construct symbolic beliefs and plan information-gathering actions. Our setting instead uses a fixed action set and an explicit belief over target identity, container relation, and presence; belief-space planning itself is not claimed as novel. Explore Until Confident~\cite{ren2024explore} and IntroPlan~\cite{liang2024introplan} use conformal prediction for stopping or instruction ambiguity; our rule uses an episode-level maximum nonconformity score for grasp eligibility in an evolving hypothesis space.

\section{Method}
\label{sec:method}

Fig.~\ref{fig:system_architecture} illustrates the overall framework and shows how categorical VLM observations update the belief used for grasp eligibility and action selection.

\subsection{Problem Formulation}

A language instruction $l$ specifies a target and, when required, its relation
to one reference container. We consider
$\mathcal R=\{\mathrm{inside},\mathrm{outside}\}$ and possible target absence.
Observation $o_t$ contains RGB-D data and camera calibration. Physical state $x_t$ contains the robot and gripper configuration, current view, and verified cover state. Let
$C_t=\{c_1,\ldots,c_{N_t}\}$ denote the set of persistent object tracks, including tracks whose objects are temporarily outside the current view. The hypothesis space is
\begin{equation}
\mathcal Z_t=
\{(c_i,r):c_i\in C_t,\ r\in\mathcal R\}
\cup\{z^{\rm unobs},z^{\rm abs}\},
\label{eq:hypothesis-space}
\end{equation}
where $z^{\rm unobs}$ denotes a present target not yet associated with a persistent track and $z^{\rm abs}$ denotes target absence. The normalized approximate belief is
\begin{equation}
b_t(z)\approx P(z_t=z\mid h_t,l),\qquad
\sum_{z\in\mathcal Z_t}b_t(z)=1,
\label{eq:joint-belief-definition}
\end{equation}
where $h_t=(o_0,a_0,\ldots,a_{t-1},o_t)$ is the history of observations and actions.
The information-gathering actions are the \emph{right} and \emph{close-high} viewpoints and cover removal when applicable. The terminal actions are grasp and defer.

\subsection{Categorical VLM Observations and Persistent Association}

Grounding DINO-Base~\cite{liu2024groundingdino} proposes candidates and
SAM~2.1 Hiera-L~\cite{ravi2025sam2} supplies masks for RGB-D localization.
Qwen3-VL-8B-Instruct~\cite{bai2025qwen3vl} evaluates each candidate through
four separate three-choice questions: color, shape, marking, and container
membership. The three identity-attribute questions use
$\{\mathrm{supported},\mathrm{contradicted},\mathrm{unobservable}\}$. The
membership question uses
$\{\mathrm{inside},\mathrm{outside},\mathrm{unknown}\}$. Identity questions use both the candidate crop and a masked image of the candidate. Container membership is assessed using the full scene.

For each three-choice question $u$ and semantic label $\lambda$, the labels are
cycled through three single-token answer letters, restored to semantic order,
and averaged at the pre-softmax logit level:
\begin{equation}
\begin{aligned}
\bar s_{t,i,u,\lambda}
&=\frac{1}{3}\sum_{m=1}^{3}
\ell_{t,i,u,m,\sigma_m(\lambda)},\\
\hat\lambda_{t,i,u}
&=\arg\max_{\lambda}\bar s_{t,i,u,\lambda}.
\end{aligned}
\label{eq:cyclic-question-score}
\end{equation}
Each semantic label occupies every answer position once. We apply neither temperature scaling nor joint logit normalization across candidate and relation hypotheses.

The four answers are deterministically combined into one of seven categorical observations,
\begin{equation}
\begin{aligned}
\mathcal Y=\{&
y_{\rm MI},y_{\rm MO},y_{\rm MU},y_{\rm mis},\\
&y_{\rm UI},y_{\rm UO},y_{\rm UU}\},
\end{aligned}
\label{eq:event-alphabet}
\end{equation}
corresponding respectively to an identity match inside the container, an identity match outside it, an identity match with unknown relation, an identity mismatch, an unknown identity inside the container, an unknown identity outside it, and unknown identity and relation. A contradicted identity attribute produces a mismatch. If all identity attributes are supported, container membership determines the match category. The remaining cases have unknown identity.

Candidate masks with IoU $\geq0.5$ against the predicted container mask are excluded. Detections with mask IoU $\geq0.75$ are grouped within each view, retaining one representative per group. Representatives are associated with persistent tracks by linear assignment using

$$
C_{\mathrm{assoc}}(i,j)
=
\frac{d(i,j)}{0.12}
+
D_{\mathrm{Hell}}(h_i,h_j),
$$

where \(d\) is the world-frame center distance in meters and \(D_{\mathrm{Hell}}\) is the Hellinger distance between 14-bin HSV histograms. The appearance term is zero when unavailable. Pairs with \(d>0.12\) m are rejected. Unmatched detections have cost 1, and matches require \(C_{\mathrm{assoc}}<1\). Unmatched representatives create new tracks, while existing tracks are retained when out of view. Association imposes no class or container-relation constraints.

\subsection{View-Conditioned Observation Model and Belief Update}
\label{sec:emission-model}

The observation model distinguishes four viewing conditions,
\begin{equation}
\begin{aligned}
\mathcal Q=\{&
\mathrm{center},\text{after removal},\\
&\text{close-high},\mathrm{right}\},
\end{aligned}
\end{equation}
and three classes
$\mathcal C=\{\text{target inside},\text{target outside},
\mathrm{distractor}\}$. For observation counts $N(q,c,y)$, the add-one-smoothed view-specific and
view-pooled observation likelihoods are
\begin{align}
E_{\rm loc}(y\mid q,c)
&=\frac{N(q,c,y)+1}{N(q,c)+|\mathcal Y|},\\
E_{\rm glob}(y\mid c)
&=\frac{N(c,y)+1}{N(c)+|\mathcal Y|}.
\end{align}
The observation likelihood used during execution is
\begin{equation}
\begin{aligned}
E(y\mid q,c)
={}&w(q,c)E_{\rm loc}(y\mid q,c)\\
&+[1-w(q,c)]E_{\rm glob}(y\mid c),\\
w(q,c)
={}&\frac{N(q,c)}{N(q,c)+1}.
\end{aligned}
\label{eq:pooled-emission}
\end{equation}
For unseen view--class pairs or views absent from the fitted model, we use the view-pooled distribution. Here $N(q,c)=\sum_y N(q,c,y)$ and
$N(c)=\sum_y N(c,y)$. No correction is applied for empirical answer-letter frequencies or priors associated with varying answer options.

After the first tracking pass, initialization uses relation weights
$\pi_{\rm R}(\mathrm{inside})=9/22$ and
$\pi_{\rm R}(\mathrm{outside})=13/22$. For $N_0>0$ tracks,
\begin{align}
b_0(z^{\rm abs})&=\frac{5}{27},&
b_0(z^{\rm unobs})&=\frac{1}{3},\\
b_0(c_i,r)&=\frac{13}{27N_0}\pi_{\rm R}(r).
\label{eq:initial-belief}
\end{align}
If no persistent track exists, $b_0(z^{\rm unobs})=22/27$ and
$b_0(z^{\rm abs})=5/27$. These priors are estimated from development-set frequencies of target presence, container relation, and track discovery using add-one smoothing.
When $m$ new tracks appear at a later step, the birth coefficient is
$\beta_{\rm birth}=(8+1)/(12+2)=9/14$ and
$\Delta_t=\beta_{\rm birth}\,b_{t-1}(z^{\rm unobs})$. Before semantic evidence is applied,
\begin{align}
b_t^-(z^{\rm unobs})
&=(1-\beta_{\rm birth})b_{t-1}(z^{\rm unobs}),\\
b_t^-(c_j,r)
&=\frac{\Delta_t}{m}\pi_{\rm R}(r),
\qquad c_j\in C_t\setminus C_{t-1}.
\label{eq:track-birth}
\end{align}
The probabilities of existing tracked-object hypotheses and $b(z^{\rm abs})$ are unchanged when new tracks are added. Without track birth, priors are retained; established tracks are not pruned by age or belief value.

For categorical observation $y_{t,i}$ from persistent track $i$, the class
implied by hypothesis $z$ is
\begin{equation}
c(z,i)=
\begin{cases}
\text{target inside}, & z=(c_i,\mathrm{inside}),\\
\text{target outside},& z=(c_i,\mathrm{outside}),\\
\mathrm{distractor},&\text{otherwise}.
\end{cases}
\label{eq:truth-class-map}
\end{equation}
The final case also applies to hypotheses for other tracks,
$z^{\rm unobs}$, and $z^{\rm abs}$. Let $\mathcal I_t$ be the distinct persistent tracks observed at step $t$, and let $n_{t,i}$ be track $i$'s observation count including the current observation. The belief update after each observation is
\begin{equation}
\begin{aligned}
b_t(z)
={}&\kappa_t\,
\max\!\left(b_t^-(z),10^{-300}\right)\\
&\times\prod_{i\in\mathcal I_t}
E(y_{t,i}\mid q_t,c(z,i))^{\,n_{t,i}^{-1/2}},
\end{aligned}
\label{eq:runtime-belief-update}
\end{equation}
with $\kappa_t$ normalizing over $\mathcal Z_t$. Observations with duplicate identifiers are counted once. Duplicate detections within a view contribute one observation per persistent track.
The $n_{t,i}^{-1/2}$ exponent tempers repeated evidence.
The tempered product of observation likelihoods approximates inference with correlated observations. It does not guarantee posterior calibration.

Because every categorical observation has the distractor likelihood under both $z^{\rm unobs}$ and $z^{\rm abs}$, a semantic update gives them
identical likelihood factors, preserving their ratio when the numerical lower bound is inactive
even though normalized masses may change. Track birth can change that ratio because it removes mass only
from $z^{\rm unobs}$. The observation model does not include a separate event for detecting no target anywhere in the scene.

\subsection{Episode-Level Conformal Commitment}

We use the split-conformal quantile construction~\cite{angelopoulos2021conformal} with $\alpha=0.1$ and $n=20$ calibration episodes. Because $\mathcal Z_t$ evolves, $z^\star_{j,t}$ is the hypothesis corresponding to the true state at time step $t$: $z^{\rm abs}$ for target absence, $z^{\rm unobs}$ before a present target is associated, and $(c^\star_{j,t},r^\star_{j,t})$ afterward. For the evaluated time steps $\mathcal T_j$, episode $j$ contributes
\begin{equation}
\rho_j=\max_{t\in\mathcal T_j}
[1-b_{j,t}(z^\star_{j,t})].
\label{eq:scene-nonconformity}
\end{equation}
With $k_\alpha=\lceil(n+1)(1-\alpha)\rceil=19$,
$\hat q_{1-\alpha}=\rho_{(k_\alpha)}$ and
\begin{equation}
\Gamma_\alpha(b_t)
=\{z\in\mathcal Z_t:1-b_t(z)\le\hat q_{1-\alpha}\}.
\label{eq:conformal-set}
\end{equation}
For the full method, \(\hat q_{0.9}=0.937\). Ablations that change the belief use separate calibration quantiles. A grasp is semantically eligible only when
$\Gamma_\alpha$ contains exactly one hypothesis specifying a candidate and container relation compatible with the instruction. Robot feasibility then checks inverse kinematics (IK), workspace and joint limits,
collision, gripper aperture, and accessibility.

Taking the maximum score within each episode avoids treating correlated time steps as independent samples. The calibration episodes were inspected during development. We therefore use the resulting threshold as a grasp eligibility rule without claiming a finite-sample sequential coverage guarantee.

\subsection{Future-Observation Model and Belief-Space Planning}

The planner uses a separate categorical model fitted from paired actions and
observations in the development data. Let $s\in\{\mathrm{covered},\mathrm{open}\}$ be
the simplified physical state. For $N_{\rm tr}>0$ known tracks, one hypothetical
future observation is approximated as
\begin{equation}
\begin{aligned}
P(s',i,y\mid z,s,a)
={}&T_a(s'\mid s)\frac{1}{N_{\rm tr}}\\
&\times E_{\rm fut}(y\mid a,y_i^{0},c(z,i)),
\end{aligned}
\label{eq:future-outcome}
\end{equation}
where $i$ is a uniformly selected known track, $y_i^{0}$ is its categorical observation at the start of planning, and $c(z,i)$ is defined by Eq.~\eqref{eq:truth-class-map}. The corresponding branch probability under belief $b$ is
\begin{equation}
\begin{aligned}
P(s',i,y\mid b,s,a)
={}&\sum_{z\in\mathcal Z_t} b(z)\\
&\times P(s',i,y\mid z,s,a).
\end{aligned}
\label{eq:future-belief-mixture}
\end{equation}
Information-gathering actions comprise cover removal and moves to the close-high or right viewpoint, each with its corresponding post-action observation model. The physical transition model uses
\begin{equation}
\begin{aligned}
T_{\rm remove}(\mathrm{open}\mid\mathrm{covered})
&=0.8,\\
T_{\rm remove}(\mathrm{covered}\mid\mathrm{covered})
&=0.2,
\end{aligned}
\label{eq:cover-transition}
\end{equation}
while an admissible viewpoint transition preserves the open state with
probability one.

The future observation model is conditioned on the action, the categorical observation at the start of planning, and the track class implied by each hypothesis. The initial observation describes appearance rather than action success or failure. For a combination represented in the fitted model,
\begin{equation}
\begin{aligned}
&E_{\rm fut}(y\mid a,y_i^0,c)\\
&\quad=
\frac{N(a,y_i^0,c,y)+E(y\mid q(a),c)}
     {N(a,y_i^0,c)+1},
\end{aligned}
\label{eq:paired-future-emission}
\end{equation}
where $q(a)$ is the viewing condition after action $a$. We estimate 15 such action-conditioned observation distributions from the development data. For an unavailable initial observation or unseen combination, the model backs off to Eq.~\eqref{eq:pooled-emission}, then to the view-pooled distribution.

The rollout samples one known track and does not model new-track discovery. Hypothetical updates omit repeated-observation tempering, and deeper nodes remain conditioned on the observation available at the start of planning.
The unobserved and absent hypotheses use distractor likelihoods; with no known track, all hypotheses receive the same likelihood and the belief is unchanged. A failed cover removal leaves the container covered but still uses the post-removal observation model. These predicted beliefs are used only to rank information-gathering actions; execution always replans from measured RGB-D.

Execution-success probabilities use a $\mathrm{Beta}(1,1)$ prior over development trial groups defined by action, seed, and scene configuration. A group is successful only if all attempts succeed, giving $p_{\rm remove}=4/5$ from $3/3$ successful groups and $p_{\rm grasp}=6/7$ from $5/5$. These fixed values model the attachment-based simulation rather than physical contact or slip and are not updated during
evaluation.

For eligible grasp hypothesis $g=(c_i,r)$,
\begin{align}
Q_{\rm grasp}(g)
&=c_{\rm act}(\mathrm{grasp})+[1-b(g)]\nonumber\\
&\quad+0.5\,b(g)[1-p_{\rm grasp}],\\
Q_{\rm defer}(b)
&=0.8[1-b(z^{\rm abs})].
\label{eq:terminal-costs}
\end{align}
The terminal-cost weights, 0.5 for the grasp execution-risk term and 0.8 for the defer cost, were empirically selected during development and held fixed for all reported evaluations.
The cost uses the joint probability $b(g)$ of the candidate and its container relation.
For a nonterminal information action,
\begin{equation}
\begin{aligned}
Q_h(b,s,a)
={}&c_{\rm act}(a)\\
&+\sum_{s',i,y}
P(s',i,y\mid b,s,a)\\
&\qquad\times
V_{h-1}\!\left(
\tau(b,s,a,s',i,y),s'
\right).
\end{aligned}
\label{eq:bellman-recursion}
\end{equation}
Here $V_h(b,s)=\min_{a'\in\mathcal A_h(b,s)}Q_h(b,s,a')$. $\mathcal A_h$ contains admissible terminal and unused information actions,
with terminal $Q_h$ given by $Q_{\rm grasp}$ or $Q_{\rm defer}$.
The operator $\tau$ multiplies belief masses by
$E_{\rm fut}(y\mid a,y_i^0,c(z,i))$ and renormalizes over $\mathcal Z_t$,
without the weighting of repeated observations used during execution. The planner uses fixed dimensionless action costs that were set during development and held constant during evaluation: 0.004 (right), 0.005 (close-high), 0.160 (cover removal), and 0.008 (grasp). Defer has zero base action cost.

The planner uses $H=3$ information-gathering action expansions followed by terminal evaluation. At depth zero, no information action is expanded. Grasp and defer are terminal and receive no future-observation cost. An eligible grasp may
terminate earlier. Reaching the horizon does not force defer when a lower-cost eligible grasp is available. Defer is admissible when no information action remains at a node or when the conformal set contains $z^{\rm abs}$ alone.
Defer means abstaining from grasping, not certifying target absence. Information actions cannot be reused within a rollout or real episode. Grasp feasibility at the current state is assessed using kinematic, collision, and gripper checks. Deeper branches use the model preconditions. The planner considers grasping at most three tracks, ranked by total belief mass, and retains the highest-belief relation for each track.

\subsection{Simulation Execution and Replanning}

In simulation, the policy replans with $H=3$ after each successful information action from the measured state and new observation. Failed interactions terminate the episode, unlike the planner's modeled continuation after failed cover removal. Thus, the predicted continuation differs from simulated execution in this case. Episodes are limited to six high-level actions. Exceeding this limit is recorded as a failure and is not treated as defer.
Measured execution outcomes change physical state but are not themselves categorical observations used in the belief update.

\section{Experiments}
\label{sec:experiments}

\subsection{Simulation Setup and Evaluation}

Simulation uses NVIDIA Isaac Sim 6.0.1 with a UR10e, RG6 gripper, and
wrist-mounted RGB-D camera. The perception system uses pretrained Grounding
DINO-Base~\cite{liu2024groundingdino}, SAM~2.1 Hiera-L
~\cite{ravi2025sam2}, and Qwen3-VL-8B-Instruct~\cite{bai2025qwen3vl}
without task-specific fine-tuning. Simulation uses an attachment-based grasp model. The resulting execution probabilities do not represent validated frictional-contact grasping.

\begin{figure*}[!t]
    \centering
    \includegraphics[width=0.95\textwidth]{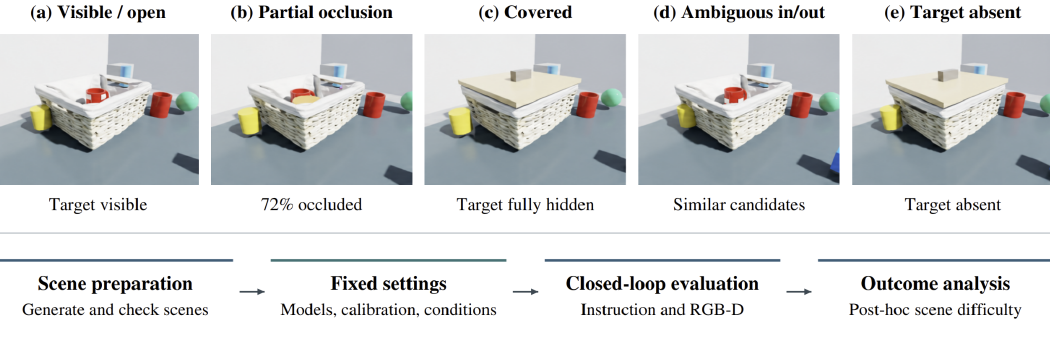}
    \caption{Representative simulated scenes (top) and evaluation protocol (bottom). Scenario labels and the displayed occlusion percentage are used only for analysis and are not provided to the policy.}
    \label{fig:scenarios_protocol}
\end{figure*}

Each method is evaluated on five randomized episodes in each of five scenarios (25 episodes total), using the same randomized instances across methods. Table~\ref{tab:evaluation_scenario_families} summarizes the main decision challenge in each scenario, and Fig.~\ref{fig:scenarios_protocol} illustrates representative scenes and the evaluation procedure. Scenario labels are used only for analysis and are not provided to the policy.

\begin{table}[t]
\centering
\caption{Scenario families and decision challenges.}
\label{tab:evaluation_scenario_families}
\footnotesize
\setlength{\tabcolsep}{3pt}
\renewcommand{\arraystretch}{1.10}
\begin{tabularx}{\columnwidth}
{@{}>{\raggedright\arraybackslash}p{0.25\columnwidth}X@{}}
\toprule
\textbf{Scenario} & \textbf{Decision challenge} \\
\midrule
Visible/Open
& Identify and retrieve the visible target without unnecessary sensing. \\
\midrule
Partially Occluded
& Acquire another view to resolve incomplete identity or visibility before commitment. \\
\midrule
Covered Container
& Remove the cover and re-observe before deciding to grasp or defer. \\
\midrule
Ambiguous Inside/Outside
& Resolve conflicting identity and container-relation evidence among similar candidates. \\
\midrule
Target Absent
& Preserve uncertainty when target evidence remains unresolved. Defer rather than commit to a distractor. \\
\bottomrule
\end{tabularx}
\end{table}

Perception evaluation uses 150 RGB images. The observation model and execution priors are estimated from separate development runs, and episode-level conformal scores use 20 calibration episodes. These calibration episodes were inspected during development. Reported evaluation outcomes are not used to update the observation model, execution probabilities, or conformal thresholds.

\subsection{Baseline Comparison and Ablation Study}

Table~\ref{tab:main_baseline_comparison} compares our proposed method with RoboEXP, DovSG,
IntroPlan, and ActPerMoMa on the same 25 scheduled evaluation episodes using a shared robot system. The adapted baselines retain the following decision mechanisms:
RoboEXP orders interactions using its scene graph and chooses the first feasible interaction. DovSG uses persistent graph memory and ordered subtasks. IntroPlan selects actions using introspective examples, with clarification requests mapped to defer. ActPerMoMa scores actions by information gain and reachability and stops according to view coverage.
All methods use shared mapping, perception, and control components for this robot. These implementations therefore do not reproduce the complete original systems. The same 25 episodes evaluated with the proposed method appear in both tables.
We selected task-adapted baselines whose decision mechanisms could be mapped to the common retrieval action set and execution interface used in our evaluation.

The ablation study evaluates the full method and eight variants, with each configuration tested on the same 25 episodes. \emph{Observations pooled across views} replaces the view-conditioned observation model with a distribution pooled across viewpoints. \emph{No cross-view memory} resets tracking, observation history, and belief after each step. \emph{Factorized belief over identity and relation} replaces the joint identity--relation belief with a factorized approximation. \emph{Hypothesis-independent future observations} uses the same predicted observation distribution across semantic hypotheses while retaining information-gathering actions and physical-state transitions.
\emph{No terminal execution-risk term} sets the modeled grasp success probability to one. \emph{No conformal grasp rule} removes only the conformal grasp-eligibility requirement, while \emph{No defer action} removes only the defer action. \emph{No conformal grasp rule or defer} removes both.

\subsection{Metrics and Statistical Analysis}

Task success is correct retrieval in target-present episodes or correct defer in target-absent episodes. False defer denotes a defer action in a target-present episode. In Tables~\ref{tab:main_baseline_comparison} and \ref{tab:ablation_study}, scenario cells report successful episodes over analyzed episodes, and aggregate task-success rates use Wilson 95\% confidence intervals. Table~\ref{tab:real_robot_evaluation} reports task success and mean episode duration over five physical-robot trials per condition. Episodes that fail before the first observation for infrastructure reasons are excluded from analyzed denominators, whereas failures after evaluation begins remain task failures.

Because DovSG and ActPerMoMa each lack one analyzed episode and the same randomized evaluation instances were reused during development, baseline rates are reported descriptively without a paired significance claim. With 25 episodes per method, results remain a controlled proof-of-concept rather than evidence for broad scenario-family generalization. Ablations are likewise interpreted descriptively because preliminary results were inspected before the final analysis protocol was fixed.

\subsection{Real-Robot Protocol}

The physical robot is a UR10e with an RG6 gripper and a wrist-mounted Zivid 2 RGB-D camera. The repeated protocol evaluates viewpoint change, cover removal, re-observation, target grasp, defer, and two injected-failure conditions, with five trials per condition. For the injected-failure conditions, the robot interface reports a prescribed failure for either the grasp or viewpoint action rather than introducing a mechanical fault. The failure status is returned to the policy, which replans autonomously without human intervention or manual trial reset. Preliminary hardware trials are excluded from this repeated evaluation. Hardware trials are separate from attachment-based simulation and are not used to fit or validate the simulator execution priors. The small number of trials demonstrates operation on the physical robot but does not provide a precise estimate of its success rate. 

\section{Results}
\label{sec:results}

\begin{table*}[!t]
\centering
\caption{Baseline task success on 25 scheduled evaluation episodes per method.
Cells show successful/analyzed episodes; DovSG and ActPerMoMa each exclude one
pre-observation timeout in the ambiguous scenario, leaving $N=24$.}
\label{tab:main_baseline_comparison}
\footnotesize
\setlength{\tabcolsep}{1.7pt}
\renewcommand{\arraystretch}{1.02}
\begin{tabular*}{\textwidth}{@{\extracolsep{\fill}}lccccccc@{}}
\toprule
Method & Analyzed $N$ & Visible & Partial & Covered & Ambig. & Absent &
\shortstack{Task success\\(Wilson 95\% CI)}\\
\midrule
RoboEXP~\cite{jiang2025roboexp} & 25 & 2/5 & 0/5 & 5/5 & 0/5 & 0/5 & 28.0\% [14.3, 47.6]\\
DovSG~\cite{yan2025dovsg} & 24 & 2/5 & 0/5 & 0/5 & 0/4 & 5/5 & 29.2\% [14.9, 49.2]\\
IntroPlan~\cite{liang2024introplan} & 25 & 5/5 & 0/5 & 0/5 & 2/5 & 5/5 & 48.0\% [30.0, 66.5]\\
ActPerMoMa~\cite{jauhri2023actpermoma} & 24 & 0/5 & 0/5 & 0/5 & 1/4 & 5/5 & 25.0\% [12.0, 44.9]\\
\midrule
\textbf{Ours} & 25 & 5/5 & 3/5 & 3/5 & 3/5 & 5/5 & \textbf{76.0\% [56.6, 88.5]}\\
\bottomrule
\end{tabular*}
\end{table*}

\begin{table*}[!t]
\centering
\caption{Ablation study on 25 episodes per variant. Cells show successful episodes out of five for each scenario family; the final three variants separately remove the conformal grasp rule, the defer action, and both.}
\label{tab:ablation_study}
\footnotesize
\setlength{\tabcolsep}{1.7pt}
\renewcommand{\arraystretch}{1.02}
\begin{tabular*}{\textwidth}{@{\extracolsep{\fill}}lcccccc@{}}
\toprule
Method variant & Visible & Partial & Covered & Ambig. & Absent &
\shortstack{Task success\\(Wilson 95\% CI)}\\
\midrule
\textbf{Ours (full)} & 5/5 & 3/5 & 3/5 & 3/5 & 5/5 & 76.0\% [56.6, 88.5]\\
Observations pooled across views & 5/5 & 2/5 & 3/5 & 3/5 & 5/5 & 72.0\% [52.4, 85.7]\\
No cross-view memory & 4/5 & 0/5 & 5/5 & 4/5 & 5/5 & 72.0\% [52.4, 85.7]\\
Factorized belief over identity and relation & 5/5 & 3/5 & 4/5 & 3/5 & 5/5 & 80.0\% [60.9, 91.1]\\
Hypothesis-independent future observations & 5/5 & 4/5 & 4/5 & 3/5 & 5/5 & 84.0\% [65.3, 93.6]\\
No terminal execution-risk term & 5/5 & 2/5 & 4/5 & 3/5 & 5/5 & 76.0\% [56.6, 88.5]\\
No conformal grasp rule & 5/5 & 3/5 & 3/5 & 3/5 & 5/5 & 76.0\% [56.6, 88.5]\\
No defer action & 5/5 & 2/5 & 3/5 & 3/5 & 0/5 & 52.0\% [33.5, 70.0]\\
No conformal grasp rule or defer & 5/5 & 4/5 & 5/5 & 3/5 & 0/5 & 68.0\% [48.4, 82.8]\\
\bottomrule
\end{tabular*}
\end{table*}

\subsection{Baseline Comparison and Ablation Results}

Table~\ref{tab:main_baseline_comparison} shows 19/25 successes (76.0\%) for the proposed method versus 12/25 (48.0\%) for the best-performing task-adapted baseline, IntroPlan. The proposed method is also the only compared policy with successes in all five scenario families. Each task-adapted baseline records zero successes in at least two scenario families.

Ablations do not produce monotonic changes in aggregate task success: factorized belief and hypothesis-independent future observations achieve 20/25 and 21/25 successes, respectively, versus 19/25 for the full method. Given the small sample, we neither interpret these variants as superior nor claim that every design choice independently improves task success.
Cross-view memory has its clearest effect under partial occlusion (3/5 to 0/5). Removing defer yields 0/5 target-absent successes, as expected because target-absent success requires abstention, while removing only the conformal grasp rule leaves binary task-success counts unchanged.

\begin{figure*}[!t]
    \centering
    \includegraphics[width=0.95\textwidth]{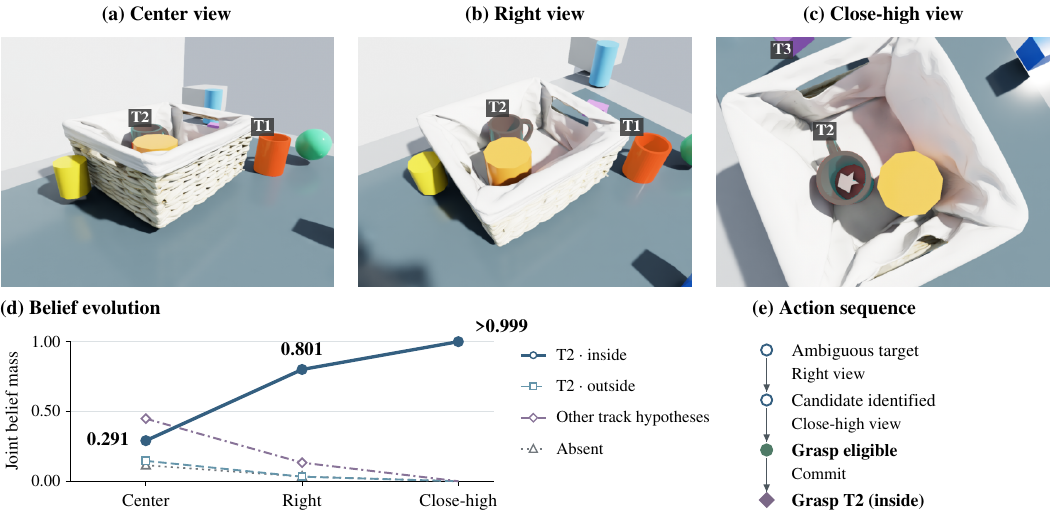}
    \caption{Example simulated retrieval episode. (a)--(c) Observations from the center, right, and close-high views. (d) The ``Other track hypotheses'' category also includes the unobserved-target hypothesis $z^{\rm unobs}$, while $z^{\rm abs}$ is shown separately. (e) The corresponding action sequence, ending with grasping T2 inside the container. ``Grasp eligible'' indicates that the conformal eligibility and robot feasibility checks are satisfied.}
    \label{fig:belief_trace}
\end{figure*}

\subsection{Qualitative Closed-Loop Trace}

Fig.~\ref{fig:belief_trace} shows a representative episode in which the policy
selects the right and then close-high viewpoints before grasping T2 inside the
container. The T2-inside belief rises from 0.291 to 0.801 and then above 0.999,
illustrating the accumulation of multi-view evidence before grasp commitment.

\subsection{Perception Frontend Evaluation}

On 150 RGB images, with operating points selected during development, Grounding DINO-Base achieved 97.4\% candidate Recall@5 and 100.0\% container-reference F1. Using the same detection boxes, SAM 2.1 Hiera-L achieved 54.4\% mask AP$_{50:95}$ and 82.7\% mean IoU. We use Grounding DINO-Base with SAM 2.1 Hiera-L in the retrieval experiments.

\begin{table}[t]
\centering
\caption{Real-robot evaluation with five trials per condition.}
\label{tab:real_robot_evaluation}
\footnotesize
\setlength{\tabcolsep}{2.0pt}
\renewcommand{\arraystretch}{1.02}
\begin{tabularx}{\columnwidth}{@{}>{\raggedright\arraybackslash}Xccc@{}}
\toprule
Condition & $N$ & \shortstack{Task\\success} &
\shortstack{Mean\\duration (s)} \\
\midrule
Visible/Open & 5 & 5/5 & 134 \\
Viewpoint change (right) & 5 & 5/5 & 160 \\
Covered Container & 5 & 5/5 & 209 \\
Target Absent & 5 & 5/5 & 175 \\
\midrule
Injected grasp-failure report & 5 & 5/5 & 180 \\
Injected viewpoint-failure report & 5 & 0/5 & 166 \\
\bottomrule
\end{tabularx}
\end{table}

\subsection{Real-Robot Evaluation}

Fig.~\ref{fig:real_robot_closed_loop} and Table~\ref{tab:real_robot_evaluation} summarize the physical-robot evaluation. The robot autonomously recovered from all five injected grasp failures, whereas all five injected viewpoint failures terminated in false defer. Mean duration ranged from 134~s for Visible/Open to 209~s for Covered Container; Table~\ref{tab:real_robot_evaluation} reports all condition-specific
durations.

\section{Discussion and Limitations}
\label{sec:discussion}

The observation model does not include a separate global non-detection event that directly updates target absence. Categorical observations can change the normalized absence probability, but $z^{\rm abs}$ and $z^{\rm unobs}$ share the same distractor likelihood in Eq.~\eqref{eq:runtime-belief-update}, so categorical observations alone do not favor target absence over unresolved target presence when the numerical lower bound is inactive. If no categorical observation or new track reaches the belief update, the belief is otherwise unchanged apart from numerical normalization; however, the episode can terminate with a perception failure when no candidate is selected before this update. The representation is also limited to one reference container, inside/outside relations, and a small set of fixed viewpoints. The future model further approximates execution by omitting new-track discovery and repeated-evidence tempering, and by simplifying failed-interaction outcomes.

The conformal rule uses only 20 calibration episodes that were inspected during development, and the execution-success probabilities come from the attachment-based simulation rather than physical contact measurements. The short planning horizon and non-reusable information actions can also exclude longer information-gathering sequences.

The ablation study does not indicate uniform additive gains. Cross-view memory has its clearest effect under partial occlusion, while the defer action is required for successful target-absent termination under the task definition; removing the conformal grasp rule alone does not change binary task-success counts. Two ablations achieve higher aggregate success than the full method, but with only 25 episodes per variant these results do not establish superiority of these variants. 

The simulation and physical-robot evaluations remain small in scope, and the baselines are task-adapted implementations rather than complete reproductions. The physical-robot study reports task success and episode duration only, so broader statistical validation will require more scenes, independent calibration data, and larger repeated hardware evaluations.

\section{Conclusion}
\label{sec:conclusion}

We present a closed-loop framework for language-guided object retrieval under partial observability. The proposed method maintains a joint belief over target identity, container relation, and presence, including separate hypotheses for an unobserved target and target absence. View-conditioned categorical VLM observations support sequential belief
updating; conformal eligibility and robot feasibility checks govern grasp commitment, while defer provides an abstention action.

Across five scenario families, our method records successful episodes in every evaluated family and achieves higher aggregate task success than the task-adapted comparison policies. The ablation study shows that individual design choices have scenario-dependent effects rather than uniform additive gains. These results support the feasibility and scenario coverage of the integrated framework, while larger evaluations are required to determine the statistical contribution of individual components. Real-robot experiments further demonstrate closed-loop interaction and autonomous recovery following injected grasp-failure reports. 

Future work will consider richer spatial relations, continuous viewpoint selection, independent calibration data, and larger-scale evaluation.

\bibliographystyle{IEEEtran}
\bibliography{references}

\end{document}